%% file: main.tex
\documentclass[10pt,twocolumn,letterpaper]{article}

\usepackage{iccv}              


\usepackage{xcolor}
\usepackage{soul}
\usepackage{threeparttable}
\usepackage{stfloats}
\usepackage{multirow}
\usepackage{array}
\usepackage{booktabs}
\newcolumntype{P}[1]{>{\centering\arraybackslash}p{#1}}
\usepackage{longtable}
\definecolor{iccvblue}{rgb}{0.21,0.49,0.74}
\definecolor{cvprblue}{rgb}{0.21,0.49,0.74}
\definecolor{lightred}{HTML}{ff9999} 
\definecolor{myorange}{HTML}{f3a977}
\definecolor{mypink}{HTML}{EC008C}
\definecolor{blueone}{HTML}{d2def1}
\definecolor{yellowone}{HTML}{fff0c2}
\definecolor{greenone}{HTML}{dcecd2}
\usepackage{pifont} 
\newcommand{\cmark}{\ding{51}}%
\definecolor{lightgreen}{HTML}{90EE90} 
\usepackage{bbding}

\definecolor{orange}{HTML}{ED8626}
\definecolor{green}{HTML}{4EA72E}
\definecolor{purple}{HTML}{7030A0}

\usepackage[pagebackref,breaklinks,colorlinks,citecolor=cvprblue]{hyperref}
\def\paperID{11144} 
\def\confName{ICCV}
\def\confYear{2025}

\title{Seeing the Trees for the Forest: Rethinking Weakly-Supervised \\Medical Visual Grounding}

\author{
Ta Duc Huy$^1$
\and Duy Anh Huynh$^1$
\and Yutong Xie $^2$
\and Yuankai Qi$^3$
\and Qi Chen$^1$
\and Phi Le Nguyen$^4$
\and Sen Kim Tran
\and Son Lam Phung$^5$
\and Anton van den Hengel$^1$
\and Zhibin Liao$^1$
\and Minh-Son To$^6$
\and Johan W. Verjans$^1$
\and Vu Minh Hieu Phan $^1$\thanks{Lead author.}
\\ 
\small{$^1$ Australian Institute for Machine Learning, University of Adelaide \qquad 
$^2$ Mohamed bin Zayed University of Artificial Intelligence}
\\
\small{$^3$ Macquarie University  \qquad $^4$ Hanoi University of Science and Technology
$^5$ University of Wollongong \qquad $^6$ Flinders University}
}

\begin{document}
\maketitle
\input{sec/0_abstract}

\input{sec/1_intro}

\input{sec/2_related}
\input{sec/3_method}
\input{sec/4_result}
\input{sec/5_conclusion}
{
    \small
    \bibliographystyle{ieeenat_fullname}
    \bibliography{main}
}
\pagebreak

\input{sec/6_supp}
\end{document}

%% file: sec/0_abstract.tex
\begin{abstract}
Visual grounding (VG) is the capability to localize specific regions in an image associated with a particular text description.
In medical imaging, VG enhances interpretability and cross-modal alignment of multimodal models by highlighting relevant pathological features corresponding to textual descriptions. 
Current vision-language models struggle to associate textual descriptions with disease regions due to inefficient attention mechanisms and a lack of fine-grained token representations. 
In this paper, we empirically demonstrate two key observations.
First, current VLMs assign high norms to background tokens, diverting the model's attention from regions of disease. Second, the global tokens used for cross-modal learning are not representative of local disease tokens. This hampers identifying correlations between the text and disease tokens. 
To address this, we introduce simple, yet effective Disease-Aware Prompting (DAP) process, which uses the explainability map of a VLM to identify the appropriate image features. 
This simple strategy amplifies disease-relevant regions while suppressing background interference. 
Without 
any additional pixel-level annotations,
DAP 
improves visual grounding accuracy by 20.74\% compared to state-of-the-art methods across three major chest X-ray datasets. 

\end{abstract}

%% file: sec/1_intro.tex
\section{Introduction}
\label{sec:intro}

Visual grounding
is a prominent research area at the vision-language intersection~\cite{deng2021transvg, dogan2019neural, karpathy2015deep}, enabling applications like visual question answering ~\cite{huang2019multi,urooj2021found}, human-robot interaction~\cite{shridhar2018interactive, tziafas2021few}, and data annotation~\cite{ichinose2023visual,wang2024learning}.
Beside concept-based approaches~\cite{Huy_2025_CVPR},
the ability of 
linking textual mentioned diseases 
to specific pathological image regions enhances interpretability~\cite{wang2024framework}, aiding radiologists in assessing model decisions.
Additionally, VG integrates effectively with large language models (LLMs) to enhance multimodal understanding, e.g., refining 3D object localization through interleaved reasoning and grounding~\cite{zhu2024scanreason}, summarizing long text with LLM to improve VG~\cite{zhu2025read}, and introducing iterative vision-language interaction with novel tokens for object detection (VG) and text generation~\cite{li2023covlm}. In summary, VG enables transparency, assists localization, and enhances multi-modal interactions on the patch level of the emerging medical multi-modal models~\cite{luo2024vividmed,chen2024chexagent,fallahpour2025medraxmedicalreasoningagent}.

Many methods~\cite{chen2023medical,ichinose2023visual,luo2024vividmed,he2024parameter} perform \textit{supervised} visual grounding on medical imaging. They rely on text-and-region correspondence annotations to train a cross-modal fusion module to ground text to specific disease regions.
However, supervised approaches require dense annotations at the pixel level, which is expensive in the medical settings.
Another work~\cite{bhalodia2021improving} proposes weakly-supervised VG for a 
pneumonia by using a pre-trained object detector. 


\begin{figure}
    \centering
    \includegraphics[width=\linewidth]
    {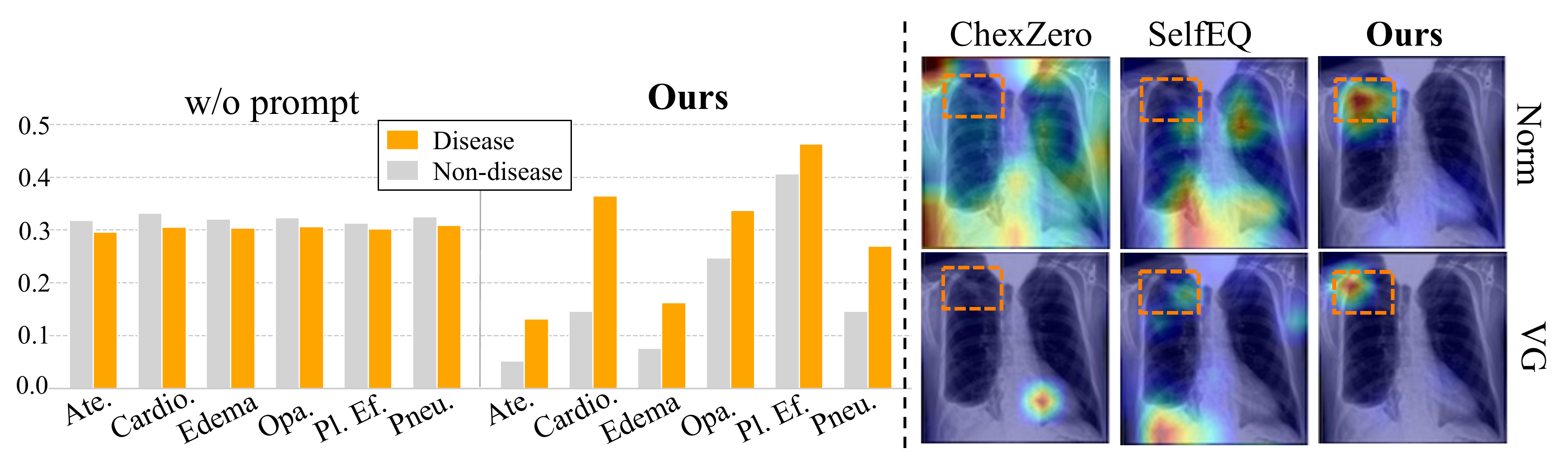}
    \vspace{-2em}\caption{\textit{Left}:
    Average cosine similarity
    between disease patch tokens and the
    corresponding disease texts of CheXzero~\cite{tiu2022expert} on MS-CXR~\cite{boecking2022making} dataset. Without disease-aware prompting, the current VLM shows undifferentiated alignment scores between disease and non-disease patches with text. \textit{Right}: Local feature norms, and visual grounding of CheXzero~\cite{tiu2022expert}, Self-EQ~\cite{he2024improved}, and our disease-aware prompted model.}
    \label{fig:teaser}
\end{figure}

Recently, weakly-supervised visual grounding methods~\cite{jiang2022pseudo,liu2023confidence,chen2018knowledge,liu2019adaptive,liu2019knowledge,rohrbach2016grounding,sun2021discriminative,liu2021relation} have leveraged strong vision-language models (VLMs) such as CLIP~\cite{radford2021learning}, as a backbone, and achieve a solid visual grounding performance without using text-to-region correspondence labels.
Most methods obtain localization pseudo-labels using explainability methods~\cite{lin2024visual,chefer2021generic} applied to the vision-language models. Alternatively, mask-driven methods train a segmentation decoder to predict the foreground and background masks of CLIP~\cite{shaharabany2022looking, shaharabany2023similarity}, or predict the masks used to mix two images~\cite{arbelle2021detector}. 
Overall, current methods get the patch-level features from the pre-trained VLM backbone on the images then aim to improve local alignment. 
Unlike general images that typically contain multiple distinct objects
, medical images typically contain more 'stuff'\cite{caesar2018coco}, and thus  present a unique challenge: critical disease indicators often occupy only a small proportion 
of the image, while anatomical backgrounds dominate visual features.
\textit{The trees are lost in the forest}. As such, applied to medical imaging, current VLMs~\cite{tiu2022expert,phan2024decomposing,bannur2023learning,boecking2022making} yield sub-optimal local features, which are less disease-focused. 


In this work, we set out to analyze this limitation in current medical VLMs, identify whether it hinders visual grounding methods, and if so to propose improvements. Our analysis shows that:
\textbf{1)} We observe that the $L_2$ norm of the token embeddings are  high  in background non-disease regions , as shown in Fig.~\ref{fig:teaser}.
%
As the norm reflects token vector magnitude and representation strength, high background norms can divert attention from disease-relevant areas, impairing medical visual grounding.
\textbf{2)} As  disease regions are typically small,  global image tokens are more strongly aligned with the major non-disease image patch tokens, and 
 less representative of the fine-grained disease tokens, as shown in Fig.~\ref{fig:problem_analysis_3}. 
As such, 
here is low alignment between disease image tokens and the global text and image tokens, which impairs visual grounding.

\begin{figure}[!h]
    \centering
    \includegraphics[width=\linewidth]{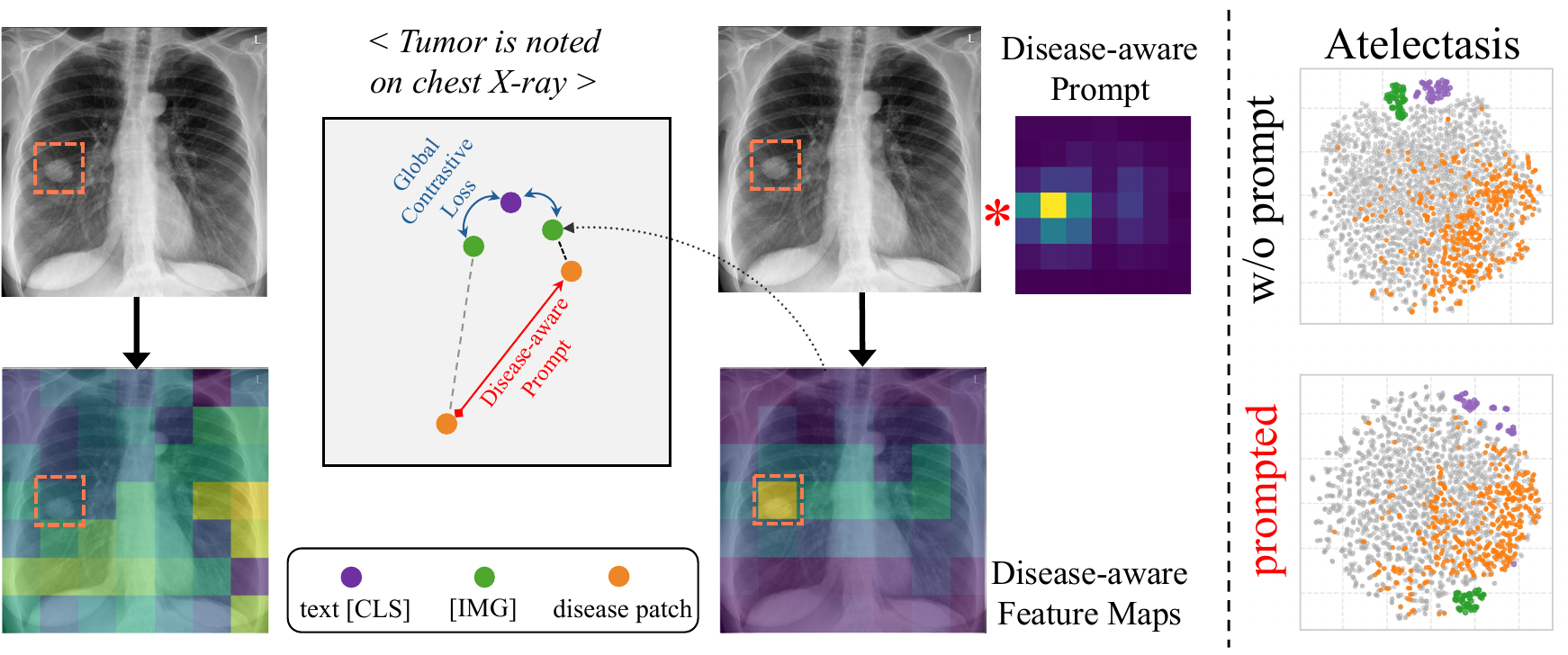}
    \vspace{-2em}
    \caption{
    In current VLMs, text [\texttt{CLS}] ({\textcolor{purple}{\scalebox{1.5}{$\bullet$}}}) are pulled close to a global image token [\texttt{IMG}] ({\textcolor{green}{\scalebox{1.5}{$\bullet$}}}) through contrastive loss, and thus they use the global token as a proxy to link with disease patch tokens ({\textcolor{orange}{\scalebox{1.5}{$\bullet$}}}). 
    \textit{Left}: w/o prompt, the global image token and the disease patch of the original input on the left show weak alignment, while our proposed \textit{disease-aware prompting} enhances contrast for disease regions, aligning disease patches with both [\texttt{IMG}] and text [\texttt{CLS}].
    \textit{Right}: The original and prompted feature space for the images having disease `atelectasis' on MS-CXR~\cite{boecking2022making} dataset. The disease tokens are pulled towards the [\texttt{IMG}] upon prompted.
    }
    \label{fig:prompt_feature}
\end{figure}


The two problems identified above illustrate  the influence of non-disease background regions on the medical visual grounding challenge. To remedy their impact,
we propose a simple-yet-effective 
\textit{disease-aware feature prompting} 
approach, as illustrated in Fig.~\ref{fig:prompt_feature}.
We first extract the explainability map~\cite{chefer2021generic} of the pre-trained VLM for the given textual disease description. We then use this disease-aware map and prompt the vision models on the feature space, thus suppressing features of non-disease tokens and amplifying disease-related tokens.  
The model then extracts a global token, \ie \texttt{[IMG]} token from this disease-focused feature map, bringing disease patches closer to the global image \texttt{[IMG]} token and improving its alignment with the text \texttt{[CLS]} token.
This design is inspired by how radiologists modify the contrast to make the disease regions more salient and localize them.
As shown in Fig.~\ref{fig:teaser} (Right), our prompting strategy effectively assigns high norm values to disease regions, producing accurate visual grounding predictions.
Without using any dense annotation, our method achieves superior performance compared to several state-of-the-art methods on multiple datasets.

Our  main  contributions are summarized as follows:
\begin{itemize}
    \item We show that medical VLMs have high norm values in non-disease background regions, which misdirects the visual grounding of fine-grained disease regions. 
    \item We further show that current VLMs suffer from \textit{intra-modal misalignment}, 
    where the global image token fails to capture the disease information from the local disease tokens.
    This  weakens local patch-to-text alignment when training via global image-text contrastive losses.  
    \item We introduce a simple-yet-effective \textit{disease-aware feature prompting} for weakly-supervised medical visual grounding. 
    Our technique suppresses the influence of non-disease local tokens to concentrate the model attention in relevant disease regions.

    \item Our proposed method achieves superior performance compared to several state-of-the-art methods for weakly-supervised medical visual grounding across three datasets. This paper is the first work, establishing a comprehensive medical weakly-supervised visual grounding benchmark on three datasets.
\end{itemize}




%% file: sec/2_related.tex
\section{Related Works}
\label{sec:related}
\textbf{Weakly-Supervised Visual Grounding} aims to localize target objects mentioned in the text without segmentation labels. 
Detector-based methods~\cite{liu2023confidence,wang2021improving,liu2021relation,gupta2020contrastive,wang2019phrase,datta2019align2ground} adopts two-stage pipeline, 
using pre-trained object detectors to extract region of interests (RoIs), and then matching texts with the cropped RoI. A recent work~\cite{liu2023confidence} generates pseudo-captions for each RoI using BLIP, and matches between the text query and the pseudo-caption.  RefCLIP~\cite{jin2023refclip}, and QueryMatch~\cite{chen2024querymatch} propose a one-stage pipeline, directly matching the text with the anchor point in the feature map or the DETR query. A state-of-the-art grounding multimodal LLM~\cite{nguyen2025localizing} uses external segmentation models to train LLMs for grounding. Unfortunately, the reliability of medical-target detectors is not comparable with the general domain. 
In contrast, detector-free methods~\cite{shaharabany2022looking,shaharabany2023similarity,arbelle2021detector} leverage pseudo labels generated by the explainability map~\cite{chefer2021generic}.
G~\cite{shaharabany2022looking} and g++~\cite{shaharabany2023similarity} re-calibrate visual feature using the image-text similarity map. They also force the model to mask the background region to yield maximum similarity with the text. G++~\cite{shaharabany2023similarity} aggregates the self-attention map having high overlap with the grounding to create a more comprehensive pseudo-label.
SelfEQ~\cite{he2024improved} encourages consistent visual grounding map with diverse text prompting. Existing WSVG methods rely on strong initial grounding map. Yet, current medical vision language models show low image-and-text alignment. 

\noindent \textbf{Vision-language models} have garnered significant interest recently, largely since the success of CLIP~\cite{radford2021learning}.
Other notable VLMs include ALIGN~\cite{jia2021scaling}, Florence~\cite{yuan2021florence}, and Flamingo~\cite{alayrac2022flamingo} have pushed forward research in this area.
Building on these advances, several medical VLMs have emerged.
These models learn the cross-modal information 
from the vision and language modalities such as chest radiographs and associated reports. CheXzero~\cite{tiu2022expert}, MedCLIP~\cite{wang2022medclip}, and BiomedCLIP~\cite{zhang2023biomedclip} align images with raw medical reports with CLIP's contrastive learning objective, while BioVIL~\cite{boecking2022making} and BioVIL-T~\cite{bannur2023learning} enhance the text encoding process.
MedKLIP~\cite{wu2023medklip} and MAVL~\cite{phan2024decomposing} leverage fine-grained report structures from RadGraph~\cite{jain2021radgraph} to further improve performance. 
MedIM~\cite{xie2024rethinking} introduces masked image modelling in medical VLMs.
Given the fine-grained nature of medical images, where diseases often occupy small areas, current medical VLMs struggle to associate thefine-grained pathological regions with textual descriptions.


\noindent\textbf{Prompt tuning} is a strategy to adapt large-scale pre-trained models to new tasks. 
CoOp~\cite{zhou2022learning} and CoCoOp~\cite{zhou2022conditional} prepend the text tokens with learnable parameters to adapt CLIP for many recognition tasks. The textual prompt tuning is widely used in open-vocabulary object detection~\cite{du2022learning} and semantic segmentation ~\cite{xu2021simple}. Several works~\cite{lin2024visual,jia2022visual,bahng2022exploring} apply visual prompt tuning on the pixel space~\cite{bahng2022exploring,xu2021simple} and feature space~\cite{jia2022visual}, to adapt a frozen pre-trained model. Previous works adopt learnable, and randomly initialized prompts for model adapting. In contrast, we adopt \textit{informative} prompting, which enforces rich disease-aware prompts to guide medical visual grounding.  

%% file: sec/3_method.tex
\section{Proposed Method}
\label{sec:method}
\noindent\textbf{Problem formulation. }
Given an input image $x_\text{v} \in \mathbb{R}^{C \times H \times W}$ and a text query $x_\text{t}$ specifying an object in $x_\text{v}$, 
weakly supervised visual grounding aims to produce a segmentation map $y \in \mathbb{R}^{H \times W}$ for the referenced object, without dense segmentation labels. 
The visual grounding model leverages a VLM with an image encoder $F_\text{v}$ and text encoder $F_\text{t}$, 
followed by a pixel decoder $D$. 
$F_\text{v}$ provides a sequence of 
local visual token features $V = \{v_1, \ldots, v_n \}$ and the global image token $\texttt{[IMG]}$ 
from $x_\text{v}$
, while $F_\text{t}$ yields 
text $\texttt{[CLS]}$ token 
from $x_\text{t}$:
\begin{align}
    \{[\texttt{IMG}], v_1, v_2, \dots, v_n\} = F_\text{v}(x_\text{v}), \quad \texttt{[CLS]} = F_\text{t}(x_\text{t}).
\end{align}

\subsection{Problem Analysis}
\label{sec:problem_analysis}
\begin{figure}
    \centering
    \includegraphics[width=\linewidth]{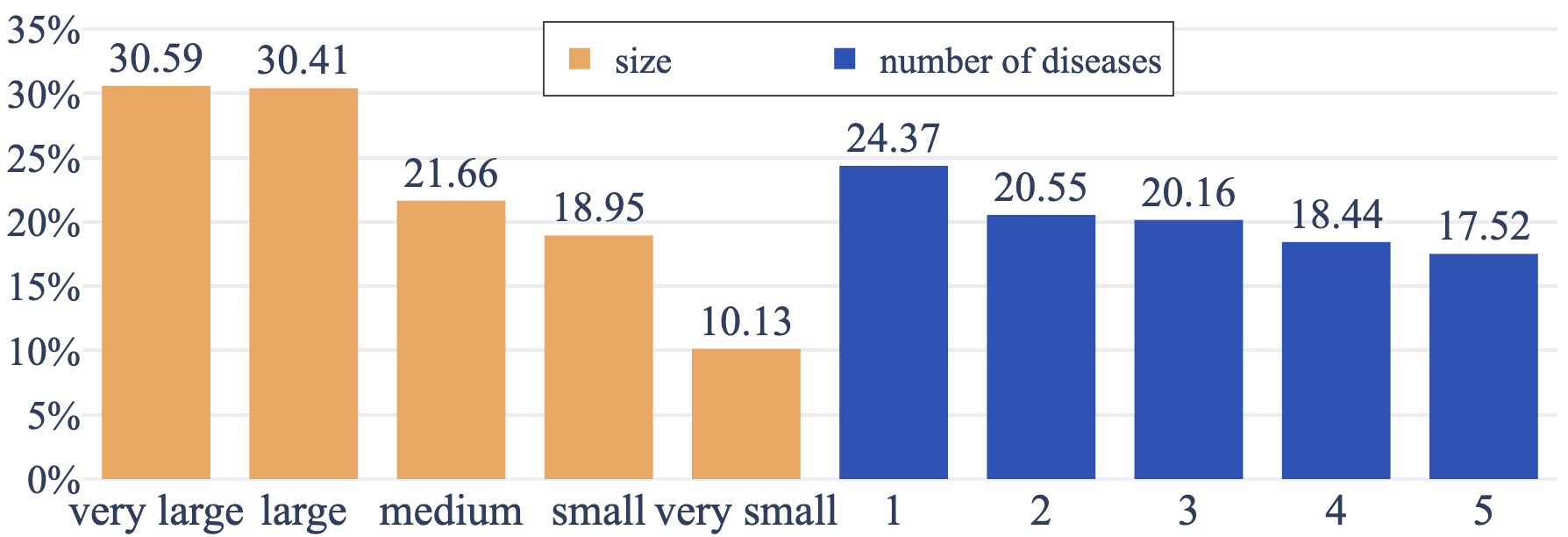}
    \caption{
    Performance analysis of BioVIL~\cite{boecking2022making} on MS-CXR dataset as a function of (a) disease region size and (b) disease count per image. \textit{Left:} Dice scores stratified by disease area, showing degraded performance for smaller regions. \textit{Right:} Performance impact of concurrent diseases, demonstrating declining accuracy with increasing disease count. Disease areas are categorized into five groups by size percentiles.
    }
    \label{fig:challenge}
\end{figure}

Current VLMs~\cite{tiu2022expert,bannur2023learning,phan2024decomposing} is efficient at identifying \textit{what} diseases are present, \ie image classification.
However, it struggles in visual grounding, determining \textit{where} to align text descriptions on the medical images. Here, we refer to the target disease regions as foreground (FG), and non-disease regions as background (BG). Fig.~\ref{fig:teaser} (Left) shows the cosine similarity distributions of text tokens with the FG and BG tokens when using the state-of-the-art CheXzero~\cite{tiu2022expert}. It shows that BG tokens have a higher alignment with the text. 
VLMs~\cite{tiu2022expert,boecking2022making,bannur2023learning} are mostly trained using only the global image token $\texttt{[IMG]}$, \ie, extracting features from the entire image and aligning with the text \texttt{[CLS]}. 
However, medical images often feature \textit{multiple} diseases with \textit{small} regions, making it difficult to correctly assign text to the corresponding regions. 
This brings challenges in fine-grained visual grounding. Fig.~\ref{fig:challenge} shows the visual grounding performance of the VL model on cases with multiple and small diseases.
The Dice drops as the disease areas reduce, and the number of present diseases increases. 

\begin{figure}[!h]
    \centering
    \includegraphics[width=1.0\linewidth]{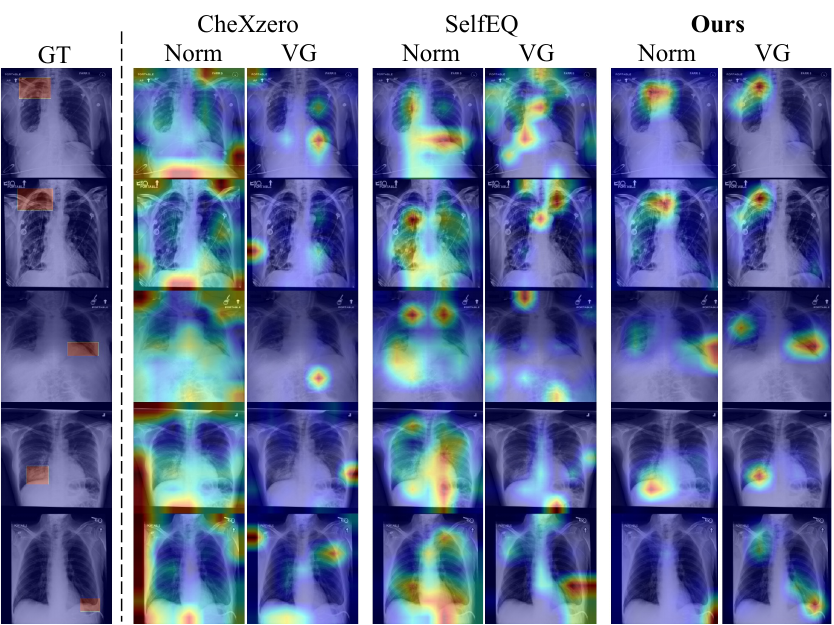}
    \vspace{-2em}
    \caption{Comparative analysis of token norm distribution (Norm) and visual grounding (VG) performance across methods with representative cases show common chest pathologies. Left to right: Ground truth (GT), CheXzero~\cite{tiu2022expert} (foundational VLM), SelfEQ~\cite{he2024improved} (weakly-supervised baseline), and our disease-aware prompting approach. 
    Our method produces more focused norm activations that better correspond to disease regions, leading to more precise grounding results. Colors indicate activation strength from low (\textcolor{blue}{blue}) to high (\textcolor{red}{red}). Text descriptions are abbreviated.}
    \label{fig:norm}
    \vspace{-1em}
\end{figure}

\noindent \textbf{High norm on BG tokens, hampering the visual grounding capability.}
The $L_2$ norm of patch tokens $\|v\|_2$ indicates the activations of the local token $v$.  Fig.~\ref{fig:norm} shows the norm of patch tokens of CheXzero. Tokens in BG regions are frequently activated with high values.
Unlike natural images, medical images are highly canonical, with significant redundancy in BG areas.
Hence, high-norm tokens often occur in these healthy BG regions rather than in the disease-associated FG areas.
%
We then show that the high activation of BG tokens hampers the visual grounding. 
Fig.~\ref{fig:norm} further compares the norm map and the visual grounding of CheXzero~\cite{tiu2022expert}. The model tends to ground on background regions with high norm.

\begin{figure}[!h]
    \centering
    \includegraphics[width=1\linewidth]{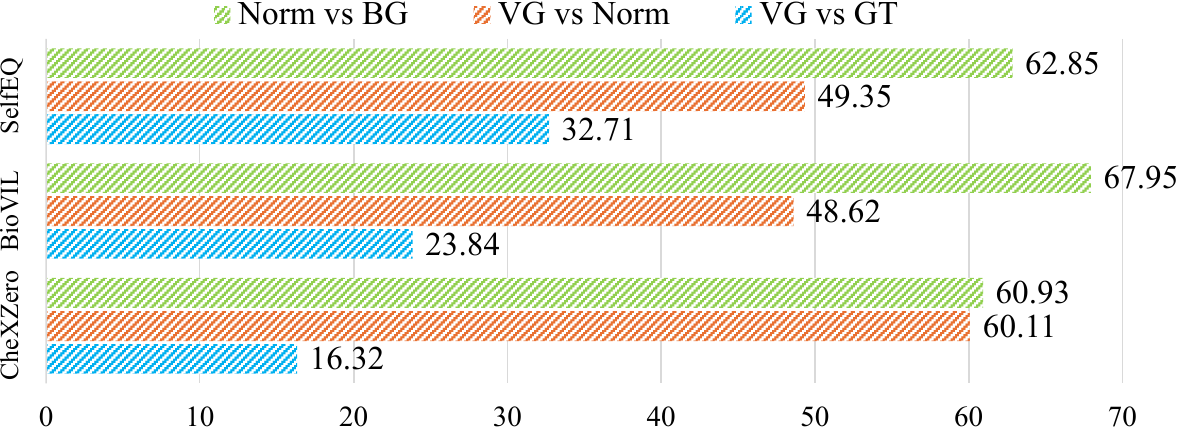}
    \vspace{-2em}
    \caption{Dice score (\%) between (i) Norm map and the background regions (BG); (ii) the visual grounding output (VG) and the norm map; and VG output and ground-truth (GT). Analysis of recent vision-language pre-training models~\cite{tiu2022expert,boecking2022making} and a recent weakly-supervised visual grounding method~\cite{he2024improved}.}
    \label{fig:norm_vs_vg}
\end{figure}

To quantify how spurious norm activation in the background (BG) regions affects visual grounding (VG), we calculate the $L_2$ norm of the tokens 
to form a norm map.
We then evaluate the correlation between these 
norm
maps and VG predictions. Fig.~\ref{fig:norm_vs_vg} shows two findings. \textit{First},  VLMs show high norm activations in the background regions, evident by the high Dice score between the response map and the BG regions. \textit{Second}, the Dice score between VG outputs on the 
norm maps
is significantly higher than the Dice with the actual ground truth (GT). 
 This shows that the models are inclined to ground on the highly activated regions of the norm map. 
%
These findings suggest that high norm activations in BG regions 
can divert the model away from disease areas, weakening VG accuracy. 

\begin{figure}
    \centering
    \includegraphics[width=\linewidth]{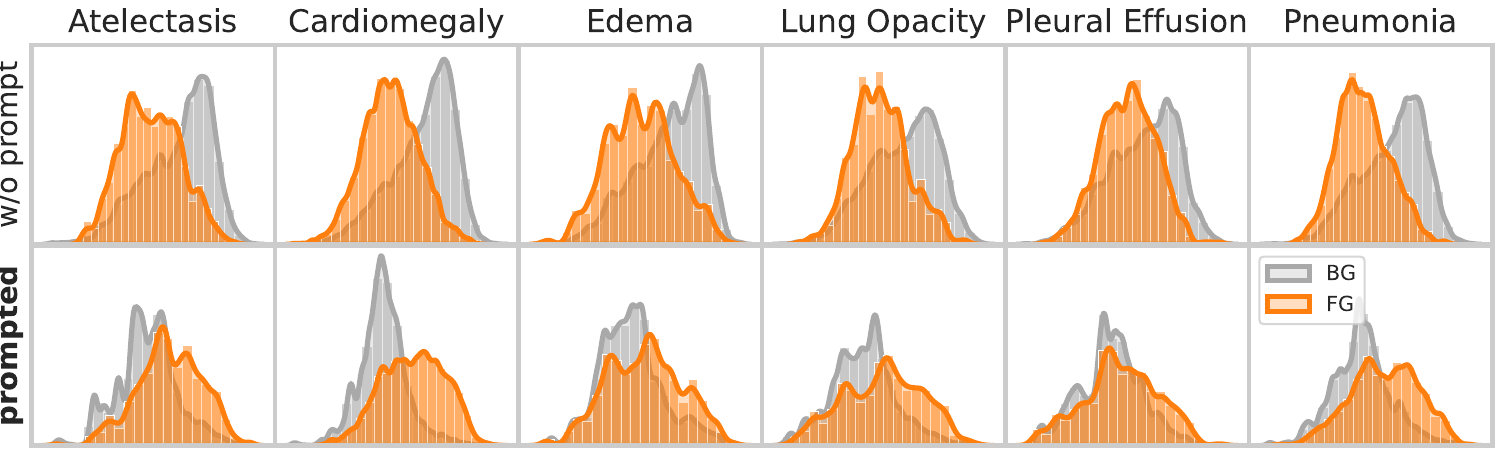}
    \caption{\textbf{Intra-modal alignment}. Distribution of cosine similarity scores between the global image $\texttt{[IMG]}$ token and the image patch tokens, for original input and prompted input. Training with the disease-aware prompts boosts the intra-modal alignment between the global token $\texttt{[IMG]}$ and disease patch tokens,  indicated by the rightward skew in the distribution of cosine similarity between \texttt{[IMG]} and FG patches.}
    \label{fig:img_alignment}
\end{figure}

\begin{figure}
    \centering
    \includegraphics[width=\linewidth]{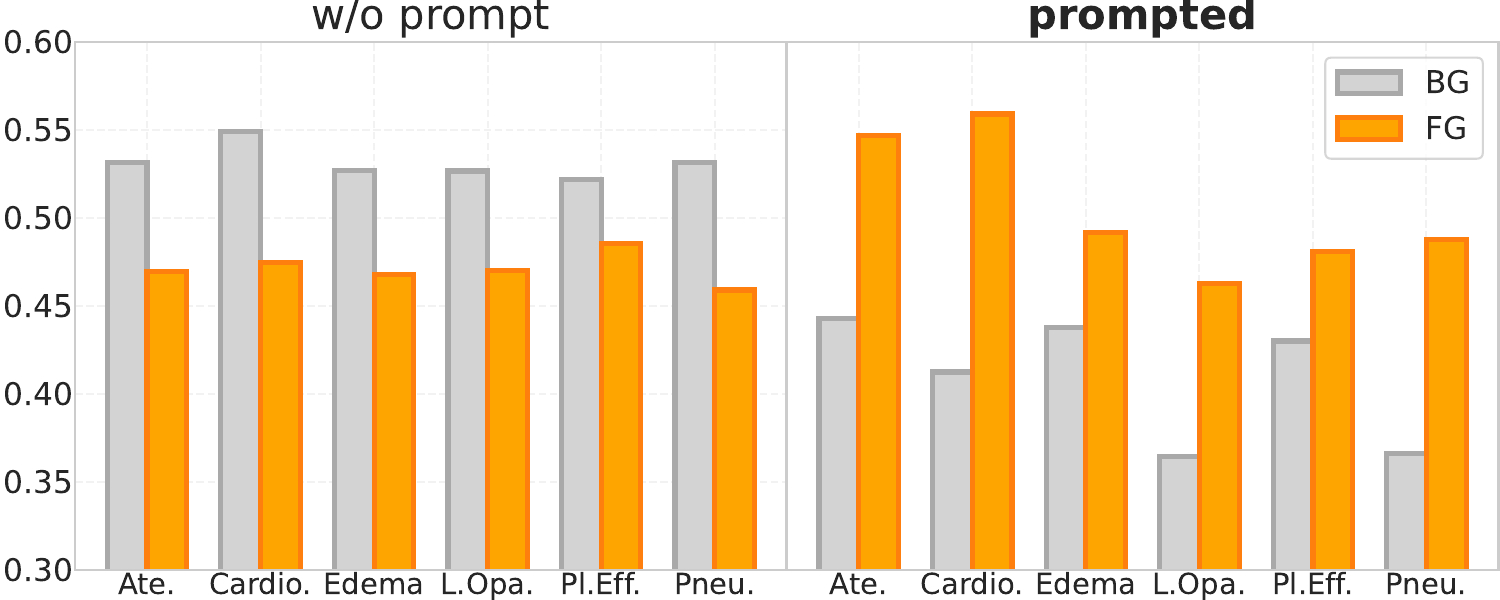}
    \caption{Average cosine similarity between the global image $\texttt{[IMG]}$ token and the image patch tokens. 
    Left: Original CheXzero model shows higher similarity to background  without prompt than disease regions. 
    Right: After prompting, the relationship inverts with stronger disease region alignment.
    }
    \label{fig:img_img_bar}
    \vspace{-1em}
\end{figure}

\noindent \textbf{Global image $\texttt{[IMG]}$ token is non-representative of the local foreground disease tokens.}
Current VLMs mostly operate on the global image $\texttt{[IMG]}$ token when aligning with the text token. Is $\texttt{[IMG]}$ representative of local patch tokens $v_i$? This study shows a negative finding.
Fig.\ref{fig:img_alignment} investigates the intra-modal alignment between the global $\texttt{[IMG]}$ token and foreground (FG) and background (BG) tokens using cosine similarity.
In the top row, the $\texttt{[IMG]}$ token aligns more closely with BG tokens than with FG tokens.
This is further substantiated by the average cosine similarity between the global $\texttt{[IMG]}$ token and individual patches (see Fig.~\ref{fig:img_img_bar}).

\begin{figure*}
    \centering
    \includegraphics[width=\linewidth]{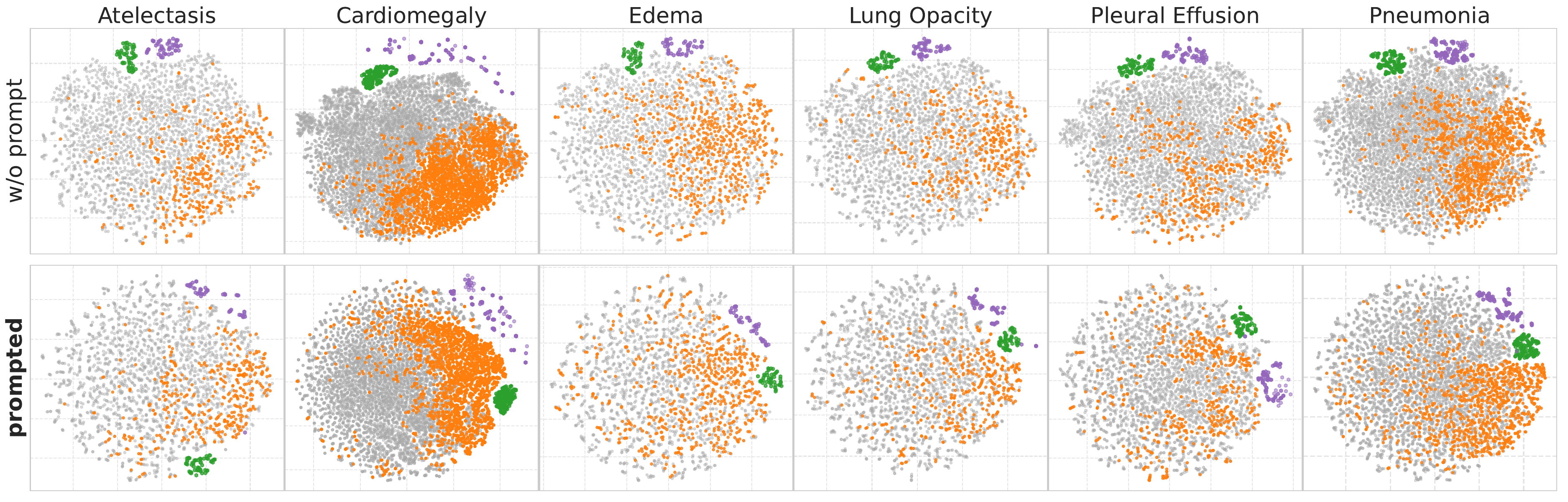}
    \vspace{-2em}
    \caption{Feature space visualization for six common chest conditions without (Top) and with our disease-focused prompting(Bottom). The global image token ({\textcolor{green}{\scalebox{1.5}{$\bullet$}}}) is close to the text token ({\textcolor{purple}{\scalebox{1.5}{$\bullet$}}}). However, the disease foreground patch ({\textcolor{orange}{\scalebox{1.5}{$\bullet$}}}) tokens are far from the global image token, thus being far from the text token. In contrast, our feature space shows higher alignment between local and global tokens, and local tokens and the text token.}
    \label{fig:problem_analysis_3}
\end{figure*}

Current VLMs use the global image $\texttt{[IMG]}$ token as an intermediary between the text $\texttt{[CLS]}$ token and the local image patches.
Hence, if the global image $\texttt{[IMG]}$ token is \textit{non-representative} of fine-grained local disease information, the alignment between the text $\texttt{[CLS]}$ and FG tokens is suboptimal. 
To validate, we show the feature distribution of global image $\texttt{[IMG]}$, text $\texttt{[CLS]}$, BG, and FG tokens in Fig.~\ref{fig:problem_analysis_3}.
Due to the global cross-modal contrastive objective, 
the image and text $\texttt{[CLS]}$ tokens are positioned closely to each other, yet both remain distant from FG disease tokens.




\begin{figure}
    \centering
    \includegraphics[width=0.97\linewidth]{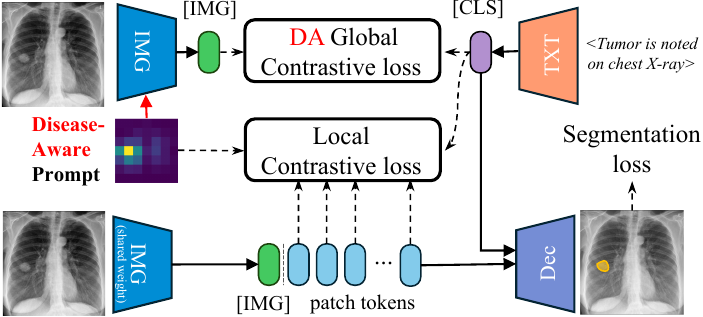}
    \caption{\textbf{Model architecture}. 
    We use the disease-aware prompt to align the global image token $\texttt{[IMG]}$ and $\texttt{[CLS]}$ for the Disease-aware global contrastive loss, and to select FG/BG patch tokens for the Local Contrastive Loss. During inference, only the original image patch token and text $\texttt{[CLS]}$ are used.}
    \vspace{-1em}
    \label{fig:method}
\end{figure}

\subsection{Disease-aware Prompting for Weakly-Supervised Visual Grounding}
\label{sec:remediation}





The study identifies two key issues: (1) weak alignment between the local tokens $V$ and global $\texttt{[CLS]}$ token, which fails to represent disease-relevant regions, and (2) high-norm background (BG) tokens that undesirably influence the visual grounding. 
As such, we propose a simple-yet-effective \textbf{disease-aware feature prompting} approach to enhance the influence of disease regions and reduce the impacts of background tokens. 
From the input image-text pair, we first extract the explainability map of the pretrained VLM.
Our framework uses this explainability map to prompt the image features, enhancing disease token contributions to the global $\texttt{[IMG]}$, suppressing irrelevant BG's influence. 
Built upon this disease-aware prompting map, we adopt a hierarchical training objective with global and local contrastive losses as illustrated in Fig.~\ref{fig:method}.

\noindent\textbf{Disease-aware feature prompting.}
To prioritize disease regions in the model’s attention, 
we apply prompting to reweight the visual token patches so that disease FG regions have stronger impact on the global image $\texttt{[IMG]}$ tokens within the Vision Transformer (ViT) via self-attention.
Employing the VLM interpretability approach~\cite{chefer2021generic}, we generate a $\text{2D}$ \textit{Disease-aware Prompt} $\Phi$ as follows:

\begin{align}
\label{eqn:gen_disease_prompt}
\Phi = \mathcal{R} \Big( V^l, \, \mathcal{M} \big( F_{\text{v}}(x_{\text{v}}), \, F_{\text{t}}(x_{\text{t}}) \big) \Big),
\end{align}
where $V^l = \{v^l_1, \ldots, v^l_n \}$ denotes the intermediate visual tokens features at the layer $l$  to interpret, $\mathcal{M}$ represents the matching function between the multi-modal outputs of the image and text encoders, and $\mathcal{R}$ applies the interpretability approach~\cite{chefer2021generic}.
Although imperfect, $\Phi$ can effectively highlight disease-relevant FG regions indicated by the text $x_\text{t}$.
We treat $\Phi$ as the disease-specific importance map for $x_\text{t}$, using it to weight the visual tokens features, thereby amplify the FG disease signals while suppressing BG noise:
\begin{align} 
    \label{eqn:prompt_mechanism} 
    \hat{V}^{l} = \Phi \cdot V^l. 
\end{align}
Accordingly, the disease-aware prompted visual output at the final layer is:
\begin{align} 
    \label{eqn:prompted_feature} 
    \texttt{[IMG]}^L_{\Phi},V^L_{\Phi} = F_\text{v}(x_\text{v}, \Phi),
\end{align}
The visual grounding output is produced by the pixel decoder, where we employed cross-attention to fuse the text and vision features:
\begin{align}
     \hat{y} = D\big(\text{CrossAttn}( V^L, \texttt{[CLS]}^L)\big).
\end{align}

\begin{figure}[!h]
    \centering
    \includegraphics[width=0.9\linewidth]{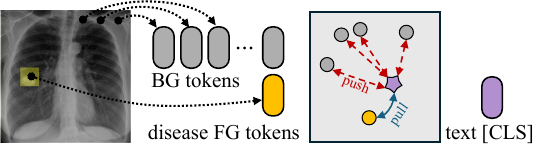}
    \caption{Local contrastive loss. The FG and BG tokens are selected using the Disease-aware prompt $\Phi$.}
    \label{fig:local_contrast}
    \vspace{-1em}
\end{figure}

\noindent\textbf{Loss objectives.} We introduce hierarchical training objective with global and local losses to maximize granularity.
First, we introduce the \textit{disease-aware global contrastive loss}. After concentrating local disease information 
into the global image $\texttt{[IMG]}$ token via disease-aware prompting, we position disease-related local tokens close to the global image $\texttt{[IMG]}$.
As such, using a global contrastive loss $\mathcal{L}_\text{glb}$, we bring the text $\texttt{[CLS]}$ closer to the image $\texttt{[IMG]}$, which serves as a \textit{proxy} to align disease-related local image tokens with the text $\texttt{[CLS]}$.
Given the batch of image-text pairs $\{ (x_\text{i}, x_\text{t})\}^n_{i=1}$, the cross-modal global contrastive loss is:
\begin{align}
    \mathcal{L}_{\text{glb}} = - \frac{1}{n} \sum_{i=1}^n \log \frac{\exp \left[ \cos\left( \texttt{[IMG]}^L_{\Phi_i}, \texttt{[CLS]}_i \right)  \right]}{\underset{j\in \{x_t^{-} \}}{\sum} \exp \left[ \cos \left( \texttt{[IMG]}^L_{\Phi_i}, \texttt{[CLS]}_j \right)  \right]},
\end{align}
where $\{x_t^{-} \}$ indicates the set of negative text descriptions of $x_i$ in the batch.

Second, we employ the local contrastive loss $\mathcal{L}_\textit{lcl}$ to align textual descriptions with disease regions in the image, as shown in Fig.~\ref{fig:local_contrast}. We threshold the disease-aware prompting mask $\Phi$ to select FG tokens $\text{FG}(V)=\{v^L\}_{\text{FG}}$ and draw them closer to the text $\texttt{[CLS]}$, while  pushing BG tokens $\text{BG}(V)=\{v^L\}_{\text{BG}}$ away. The local contrastive loss for each image-text pair is defined as: 
\begin{align}
\mathcal{L}_{\text{lcl}} = \nonumber
- \frac{1}{|\text{FG}_i|} 
\sum_{v_{\text{fg}} \in \text{FG}_i}
\log \frac{\exp \left[ \cos\left( v_{\text{fg}}, \texttt{[CLS]}_i \right) \right]}{
\sum\limits_{v_{\text{bg}} \in \text{BG}_i} \exp \left[ \cos \left( v_{\text{bg}}, \texttt{[CLS]}_i \right) \right]}.
\end{align}

\begin{figure*}[!h]
    \centering
    \includegraphics[width=0.93\linewidth]{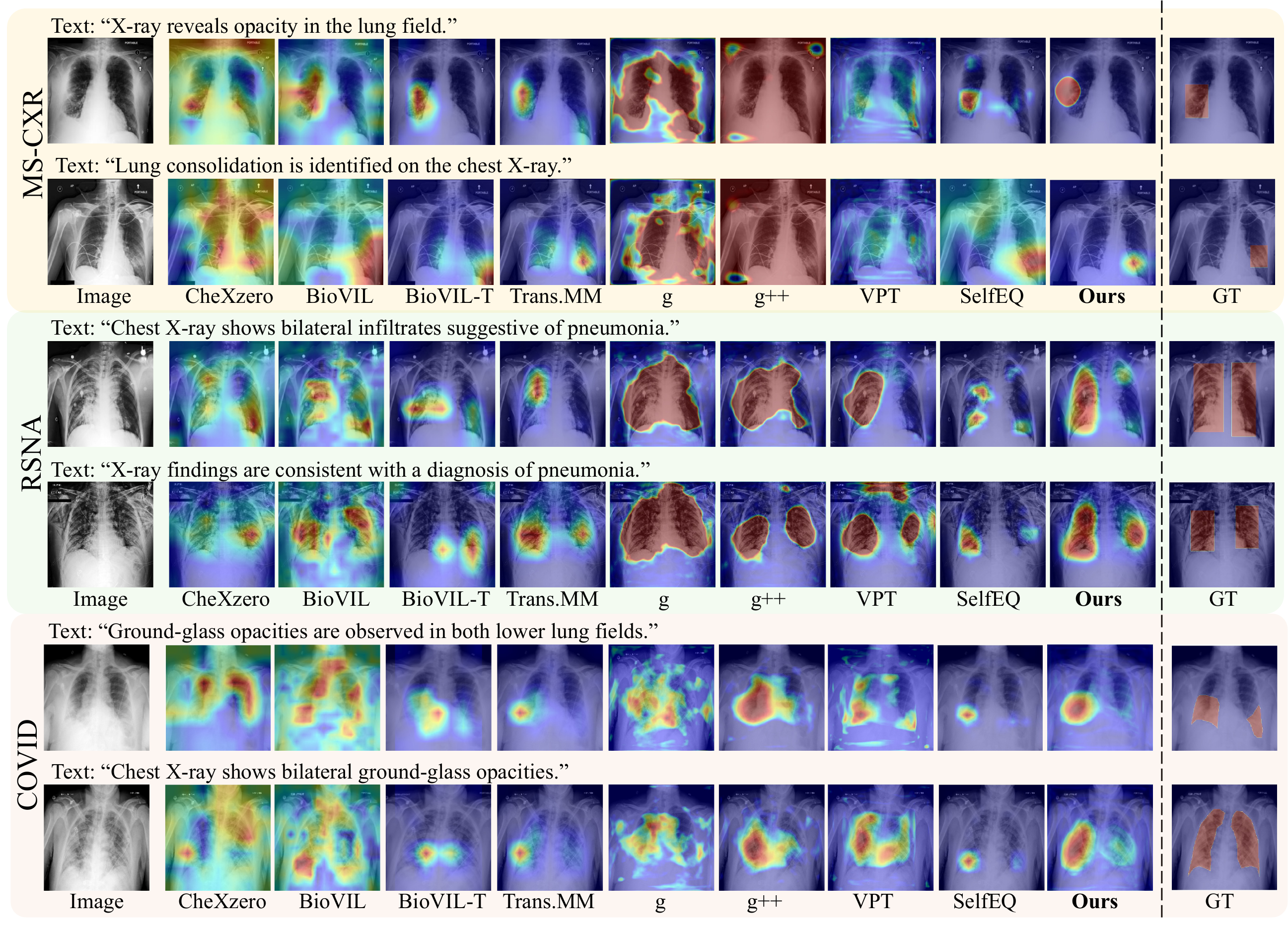}
    \vspace{-1.1em}
    \caption{Qualitative results of recent VLMs, and weakly-supervised VG methods on MS-CXR\cite{boecking2022making} , RSNA~\cite{shih2019augmenting} and Covid19~\cite{desai2020chest}.}
    \label{fig:quali_ms_rs}
\end{figure*}

Finally, we learn the pixel decoder with $\Phi$ as pseudo-label, using Dice loss $\mathcal{L}_\text{seg}$.
The total loss objective is:
\begin{align}
    \mathcal{L} = \mathcal{L}_\text{glb} + \mathcal{L}_\text{lcl} + \mathcal{L}_\text{seg}
\end{align}

Note that, the extracted disease-aware prompt $\Phi$ is only needed during training to resolve the local misalignment of VLM. Our disease-aware contrastive loss $\mathcal{L}_{\text{glb}}$ pulls the text $\texttt{[CLS]}$ close to the more disease-representative global token $\texttt{[IMG]}$, thus improving the local alignment between text and local disease tokens $V$. Hence, during inference, our model can ground disease patches on the original image without relying on the pre-extracted explainability map. Using this map to narrow down the \textit{pathological search space}, our DAP enables the model to identify subtle and fine-grained disease features, thus further improving the disease visual grounding. 



\noindent\textbf{Discussion.}
While a recent supervised visual grounding method, ZSeg~\cite{xu2021simple}, also proposes to mask the input image, our disease-aware prompting differs.
ZSeg masks in the pixel space; 
in contrast, we propose to prompt in the \textit{feature} space to address the followings. 
First, pixel hard-masking can erase relevant context, while we multiply $\Phi$ with token features, preserves both local and global information.
Second, pixel masking is insufficient to suppress high background norms as it only operates at the image level. 
Our method effectively reduces high background norm compared to pixel masking as shown in Fig.~\ref{fig:pixel_mask}.


%% file: sec/4_result.tex
\begin{table*}[!h]
\centering
\caption{Weakly-supervised visual grounding benchmark of different methods on MS-CXR~\cite{boecking2022making} and RSNA~\cite{shih2019augmenting} on CNR as our main metric, and other standard visual grounding metrics.}
\label{tab:benchmark}
\resizebox{0.8\textwidth}{!}{
\begin{tabular}{cc|cccc|cccc}
     \toprule \textbf{Dataset} & & \multicolumn{4}{c|}{\textbf{MS-CXR}} & \multicolumn{4}{c}{\textbf{RSNA Pneumonia}}  \\\cmidrule{3-6} \cmidrule{7-10}
     \textbf{Method} & Venue & CNR$\uparrow$ & PG $\uparrow$ & Dice $\uparrow$  & IoU $\uparrow$ & CNR $\uparrow$ & PG $\uparrow$ & Dice $\uparrow$ & IoU  $\uparrow$\\\midrule
     CheXzero~\cite{tiu2022expert} & Nat. BE 22 & 0.675 & 0.156 & 0.163 & 0.089 & 0.814 & 0.468 & 0.202 & 0.120    \\
     BioViL~\cite{boecking2022making} & ECCV22 & 0.941 & 0.326 & 0.238  & 0.216 & 1.096 & 0.472 & 0.339 & 0.252  \\
     BioViL-T~\cite{bannur2023learning} & CVPR23 & 1.164 & 0.416 & 0.320 & 0.218 & 0.805 & 0.584 & 0.295 & 0.192 \\\midrule
     Trans-MM~\cite{chefer2021generic} & ICCV21  &1.122 & 0.406 & 0.336 & 0.244 & 1.293 & 0.526 & 0.421 & 0.284 \\
     g~\cite{shaharabany2022looking} & NIPS22 & 0.672 & 0.304 & 0.290 &  0.186 & 1.461 & 0.284 & 0.450  & 0.315 \\
     g++~\cite{shaharabany2023similarity} & CVPR23 & 0.569 & 0.258 & 0.268 & 0.152 & 1.350 & 0.467 & 0.445 & 0.311  \\
     VPT~\cite{lin2024visual} & ICASSP24 & 0.588 & 0.323 & 0.255 & 0.162 & 1.173 & 0.612 & 0.468 & 0.327 \\
     Self-EQ~\cite{he2024improved} & CVPR24 & 1.082 & 0.406 & 0.327 & 0.229 & 1.075 & 0.637 & 0.370 & 0.241  \\\midrule
     DAP  & Proposed & \textbf{1.254} & \textbf{0.457} & \textbf{0.352} &  \textbf{0.245}& \textbf{1.630} & \textbf{0.747} & \textbf{0.474} & \textbf{0.328} \\\bottomrule
\end{tabular}
}%
\vspace{-1em}
\end{table*}

\section{Experimental settings}
\label{sec:results}
\textbf{Datasets.}  MIMIC-CXR v2 chest radiograph data~\cite{johnson2019mimic} comprises 227,835 imaging studies from 65,379 patients with image-level labels. 
We include only frontal-view scans, 
utilizing in a train/validation set of 146.7k/22.2k samples.
Evaluation was done using MS-CXR~\cite{boecking2022making}, RSNA Pneumonia~\cite{shih2019augmenting} and COVID Rural~\cite{desai2020chest} datasets.
MS-CXR~\cite{boecking2022making} is a subset of MIMIC, provides bounding boxes for 1162 images on 8 diseases. 
RSNA Pneumonia~\cite{shih2019augmenting} includes 26k frontal chest X-rays with pneumonia masks, provided by the Radiological Society of North America. 
COVID Rural~\cite{desai2020chest} contains over 200 annotated chest X-rays for COVID-19 segmentation.
We use GPT4-o~\cite{openai2023gpt4} to generate 50 unique text descriptions per disease label, which are reviewed by board-certified radiologists to create a refined text description database. 
During training, models are prompted with a random sample from its respective disease descriptions set.


\noindent \textbf{Comparison approaches.} We benchmark two groups. The first group is the medical VL pre-training models that are trained on a large-scale image-report pairs, including BioViL~\cite{boecking2022making}, BioViL-T~\cite{bannur2023learning}, and CheXzero~\cite{tiu2022expert}. The second group includes state-of-the-art weakly visual grounding models, which are trained via the pseudo-labels generated from the pre-trained VLMs. These include Self-EQ~\cite{he2024improved}, g~\cite{shaharabany2022looking}, g++~\cite{shaharabany2023similarity}, Transformer-MM~\cite{chefer2021generic}, and VPT~\cite{lin2024visual}. 
Since medical images do not have a strong and universal disease object detectors, we only benchmark against detector-free VG methods, and exclude detector-based methods~\cite{jiang2022pseudo,liu2021relation}.

\noindent \textbf{Metrics.} Following recent medical visual grounding methods~\cite{boecking2022making,bannur2023learning}, we adopt the contrast-to-noise ratio (CNR) as the main metric, which measures the difference between FG and BG activations.
Besides, we adopt standard VG metrics, including pointing game (PG), Dice and IoU scores.

\noindent \textbf{Implementation Details.} Our experiments are conducted on 4 NVIDIA RTX A100 GPUs. Input images are resized to 224$\times$224 and augmented with RandAugment~\cite{cubuk2020randaugment}.
We set up an Adam optimizer~\cite{diederik2014adam} with a learning rate of 0.008 and a batch size of 512 across all experiments.
We cap the maximum length of each text to 77 as suggested by~\cite{tiu2022expert}. 
Our hyperparameter values and schedules were determined empirically on a small validation subset.


\section{Results and Discussions}
\textbf{Comparisons with state-of-the-art methods.} Table~\ref{tab:benchmark} shows the results of recent methods on the two datasets. Our method demonstrates a notable improvement over the vision-language model BioViL, achieving a CNR boost of 7.73\% and 48.72\% respectively on MS-CXR and RSNA datasets. 
When compared with Self-EQ~\cite{he2024improved}, a recent advancement in weakly-supervised visual grounding, our approach significantly outperforms it by 15.89\% in CNR and 9.36\% in the pointing game on MS-CXR. Furthermore, we surpass g++~\cite{shaharabany2023similarity} by 20.74\% in CNR on the RSNA dataset.

Table~\ref{tab:covid} presents the comparisons on the Covid-19 segmentation dataset. We significantly outperforms the recent weakly-visual grounding methods, VPT, and Self-EQ by 17.7\% and 33.26\% in terms of PG.
Fig.~\ref{fig:quali_ms_rs} shows that existing visual grounding methods tends to over-segment the disease regions. These results underscore that existing weakly-supervised visual grounding methods fall short on medical imaging tasks, mainly due to the limited disease-focused local features within the VLM backbone. In contrast, ours show crisp localization on the fine-grained disease regions.

\begin{table}
\centering
\caption{Weakly-supervised visual grounding benchmark of different methods on Covid19~\cite{desai2020chest}.}
\label{tab:covid}
\resizebox{0.46\textwidth}{!}{
\begin{tabular}{cc|cccc}
     \toprule \textbf{Method} & Venue & CNR$\uparrow$ & PG $\uparrow$ & Dice $\uparrow$ & IoU $\uparrow$ \\\midrule
     CheXzero~\cite{tiu2022expert} & Nat. BE 22 & 0.923 & 0.235 & 0.165 & 0.097    \\
     BioViL~\cite{boecking2022making} & ECCV22 & 0.623 & 0.438 & 0.177 & 0.201 \\
     BioViL-T~\cite{bannur2023learning} & CVPR23 & 0.739 & 0.469 & 0.297  & 0.220 \\\midrule
     Trans-MM~\cite{chefer2021generic} & ICCV21  & 0.955 & 0.406 & 0.325 & 0.223  \\
     g~\cite{shaharabany2022looking} & NIPS22 & 0.610 & 0.282 & 0.246 & 0.157  \\
     g++~\cite{shaharabany2023similarity} & CVPR23 & 0.702 & 0.469 & 0.311 & 0.205  \\
     VPT~\cite{lin2024visual} & ICASSP24 & 0.739 &  0.531 & 0.332 & 0.220 \\
     Self-EQ~\cite{he2024improved} & CVPR24 & 0.659 & 0.469 & 0.304 & 0.190  \\\midrule
     DAP  & Proposed & \textbf{1.018} & \textbf{0.625} & \textbf{0.391} &  \textbf{0.267} \\\bottomrule
\end{tabular}
}%
\vspace{-0.5em}
\end{table}

\begin{figure}
    \centering
   
    \includegraphics[width=1\linewidth]{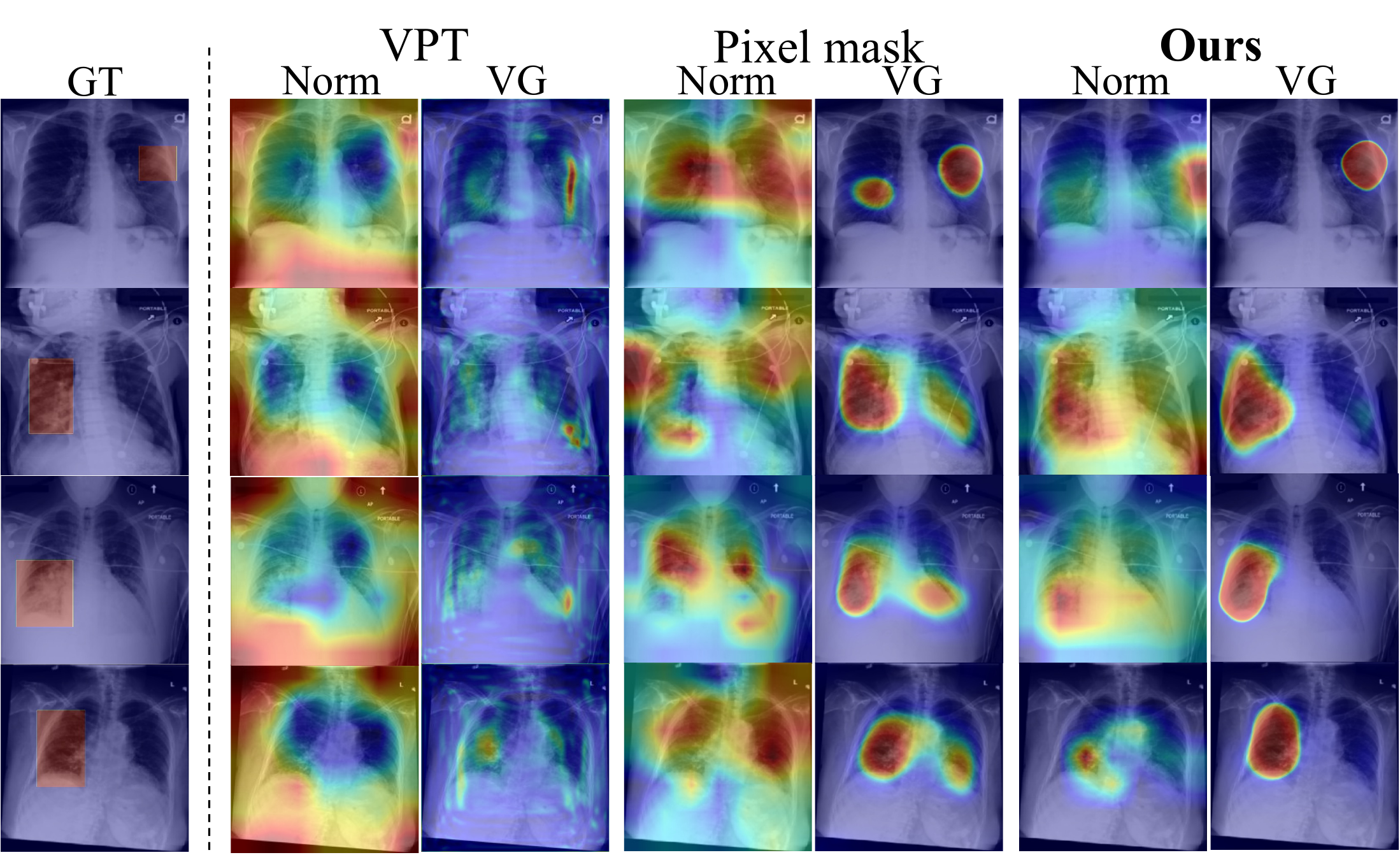}
     \vspace{-2em}
    \caption{Token norm and VG analysis of prompt tuning methods: VPT~\cite{jia2022visual}, Pixel mask~\cite{xu2021simple} versus our DAP.}
    \label{fig:pixel_mask}
    \vspace{-1.5em}
\end{figure}

\noindent \textbf{Ablation Study.} \noindent \textit{Norm and visual grounding of prompt tuning methods.}
We evaluate the token norms of visual prompts in VPT~\cite{lin2024visual} and pixel prompts in 
ZSeg~\cite{xu2021simple}.VPT only learns the prompt and does not overcome the problem of high norms in BG regions.
Pixel prompting reduces BG noise at the pixel level, but its effect weakens in deeper layers.
Our approach, which prompts directly in the feature space, provides consistent background suppression, with improved qualitative results demonstrated in Fig.~\ref{fig:pixel_mask}.

\noindent \textit{Comparisons of different prompting.} We compare VG performance of different  prompting techniques, as shown in Table~\ref{tab:prompt}. 
Notably, we compares ours with pixel prompting adopted in 
ZSeg~\cite{xu2021simple},
which applies object binary masks on the image pixels. Ours significantly improves upon the pixel masking prompting by 32\% in CNR metrics. 
Ours explicitly suppresses the non-disease regions, thus optimally alleviating the influence of background regions.
Secondly, our disease-aware prompting outperforms learnable textual prompts (e.g., CoCoOp~\cite{zhou2022learning}), visual prompts~\cite{jia2022visual}, and multi-modal prompts~\cite{khattak2023maple}.
This shows that \textit{informative} disease-aware prompting is more suitable for fine-grained medical visual grounding. 
\begin{table}[!]
    \centering
    \caption{Comparisons of different prompt tuning techniques applied on weakly-supervised visual grounding on the MS-CXR dataset~\cite{boecking2022making}.}%
    \label{tab:prompt}
    \resizebox{0.47\textwidth}{!}{
    \begin{tabular}{c|cccc}
         \toprule Prompting & CNR$\uparrow$ & PG$\uparrow$ & Dice$\uparrow$ & IoU$\uparrow$  \\\midrule
         Red circle~\cite{shtedritski2023does} & 0.866 & 0.307 & 0.261 & 0.177 \\
         Pixel mask~\cite{xu2021simple} & 0.949 & 0.361 & 0.306 & 0.199 \\
        CoOp~\cite{zhou2022learning} & 0.986 & 0.364 & 0.311 & 0.206 \\
         CoCoOp~\cite{zhou2022conditional} & 1.042 & 0.418 & 0.333 & 0.226 \\
         Visual prompt~\cite{jia2022visual} &  1.129 & 0.418 & 0.338 & 0.231 \\
         Multi-modal~\cite{khattak2023maple} & 1.101 & 0.414 & 0.345 & 0.233 \\\midrule 
         Disease-aware (Ours) & \textbf{1.254} & \textbf{0.457} & \textbf{0.352} & \textbf{0.245}  \\\bottomrule
    \end{tabular}%
    }
\end{table}

\noindent \textit{Ablation study of the proposed components.} Table~\ref{tab:abl} shows the performance when ablating different components. First, adding disease-aware prompting into the global contrastive loss significantly improves the Dice score by 24.01\%. This shows that using VLM with only global contrastive loss can lead to the low local alignment, hindering the visual grounding. Second,  integrating DAP and global contrastive loss yields comparable performance with the local contrastive loss. Without fine-grained local objective, our disease-aware prompting helps pull the local patches closer to $\texttt{[CLS]}$ only using the global token $\texttt{[IMG]}$ as a proxy. Applying all components yield the highest performance, showing the complementary of disease-aware prompting and global-to-local objectives for fine-grained alignment.

\begin{table}[!]
    \centering
    \caption{Ablation study of the disease-aware prompting (DAP), the global contrastive loss, and the local contrastive loss on the MS-CXR dataset.}
    \label{tab:abl}
    \begin{tabular}{ccc|cccc}
         \toprule DP & $\mathcal{L}_\text{glb}$  &  $\mathcal{L}_\text{lcl}$ & CNR$\uparrow$ & PG$\uparrow$ & Dice$\uparrow$ & IoU$\uparrow$  \\\midrule
          & \cmark &  &0.878 & 0.376 & 0.279 & 0.153\\
          \cmark & \cmark & & 1.061 & \underline{0.441} & \underline{0.346} & 0.237 \\
          &  &  \cmark & \underline{1.064} & 0.434 & 0.343 & \underline{0.238}  \\
          \cmark & \cmark & \cmark & \textbf{1.254} & \textbf{0.457} & \textbf{0.352} & \textbf{0.245} \\\bottomrule
    \end{tabular}
\vspace{-1em}
\end{table}

%% file: sec/5_conclusion.tex
\section{Conclusion}
\label{sec:conclusion}
This paper provides a diagnostic view of local alignment of 
medical VLM, hampering the weakly-supervised visual grounding. We show that (i) these models assign high norm values to non-disease regions, misdirecting attention, and (ii) intra-modal misalignment between global and local tokens weakens disease-specific grounding.
As such, we propose 
DAP
that strengthens disease-region representations while suppressing background interference. 
This paper constructs a comprehensive weakly-supervised medical VG on three datasets to catalyze the research, where DAP 
 ours outperforms state-of-the-art methods by up to 20.74\%.

%% file: sec/6_supp.tex




\definecolor{cvprblue}{rgb}{0.21,0.49,0.74}



\def\paperID{11144} 
\def\confName{ICCV}
\def\confYear{2025}



\thispagestyle{empty}
\appendix

\section{Self-enhancement with DAP}
\begin{figure}[h]
    \centering
    \includegraphics[width=1\linewidth]{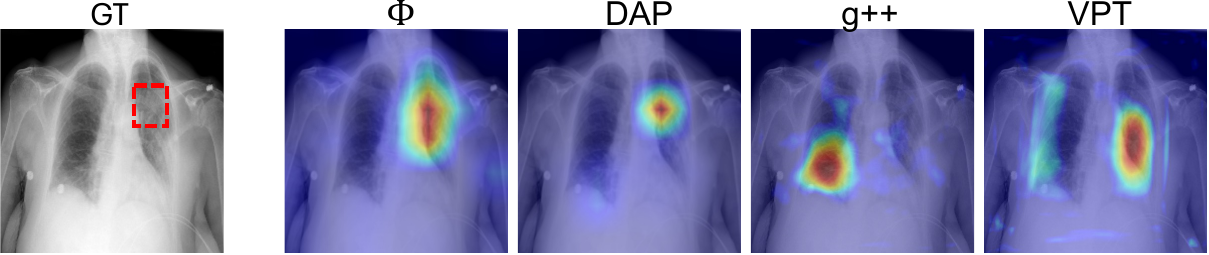}
    \caption{Results of interpretability map, our DAP, and others.}
    \label{fig:c}
\end{figure}

The interpretability map $\Phi$ shows a reasonable localization capability as it achieves 33.6\% average dice score on MS-CXR.
Furthermore, $\Phi$ helps enhance the model by narrowing down the pathological search space for VG.
By dampening the influence of background, the model can discover finer pathological signals, which get dominated without the initial interpretability map prompting.
Fig.~\ref{fig:c} shows that our model further refines the localization ofter $\Phi$ narrows down the search space.

\begin{figure}[h]
    \centering
    \includegraphics[width=1\linewidth]{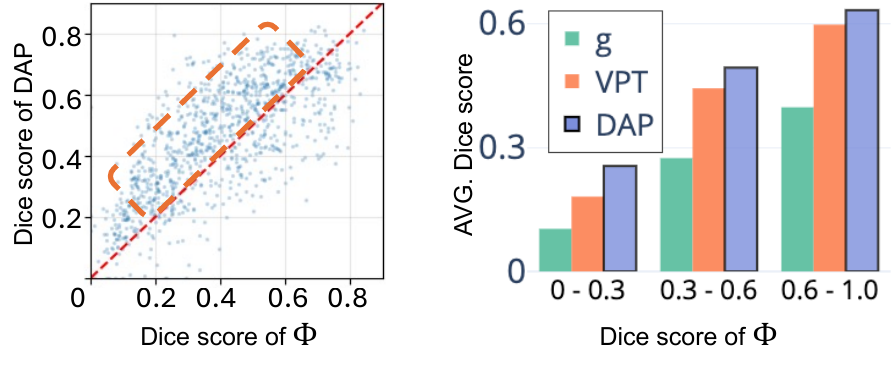}
    \caption{\textit{Left:} Relationship between the performance of the interpretability map and our DAP. Points above the red line (highlighted in \textcolor{orange}{orange}) indicate \textit{self-enhancement} cases, where our DAP further improves the Dice score of interpretability maps. \textit{Right:} Avg. Dice on samples grouped by the quality of the interpretability map. Dice of other VG are shown for comparisons.}
    \label{fig:rebuttal}
\end{figure}

Fig.~\ref{fig:rebuttal} (Left) plots the Dice score of our model against $\Phi$ on RSNA. Most points lie above the red line, showing our model imprives Dice upon the interpretability map, indicating self-enhancement.
Fig.~\ref{fig:rebuttal} (Right) shows DAP surpasses baselines even when $\Phi$ fails (Dice $<$ 0.3).

\section{Hyperparameters tuning}
In this section, we study the impact of different hyperparameter settings, including the batch size, learning rate, the weights of loss objectives, text prompts variety and prompt depth. We conduct the hyperparameters tuning on 20\% of the MS-CXR~\cite{boecking2022making} dataset.
\subsection{Batch Size and Learning Rate}
We vary the batch size of $\{128, 256, 512\}$ and report the result in Table.~\ref{tab:batch_size}.
It is shown that larger batch size has a positive impact on the overall performance.

\begin{table}[h]
    \centering
    \caption{Batch size versus performance.}
    \label{tab:batch_size}
    \begin{tabular}{c|ccc}
         \toprule batch size & CNR$\uparrow$ & PG$\uparrow$ & Dice$\uparrow$  \\\midrule
          128 & 1.027 & 0.443 & 0.340 \\
          256 & 1.038 & 0.440 & 0.347  \\
          512 & 1.042 & 0.449 & 0.350 
          \\\bottomrule
    \end{tabular}
\end{table}

We vary the learning rate of $\{1e^{-1}, 1e^{-2}, 1e^{-3}\}$ and report the result in Table.~\ref{tab:learning_rate}. The optimal learning rate is $1e^{-3}$.
We set batch size to 512 and learning rate to $1e^{-3}$ for other experiments.

\begin{table}[h]
    \centering
    \caption{Learning Rate versus performance.}
    \label{tab:learning_rate}
    \begin{tabular}{c|ccc}
         \toprule lr & CNR$\uparrow$ & PG$\uparrow$ & Dice$\uparrow$  \\\midrule
          $1e^{-1}$ & 0.917 & 0.358 & 0.288 \\
          $1e^{-2}$ & 1.054 & 0.425 & 0.345  \\
          $1e^{-3}$ & 1.042 & 0.449 & 0.350 \\\bottomrule
    \end{tabular}
\end{table}

\subsection{Loss weights}

We vary the weights of the Disease-aware Global contrastive loss $\mathcal{L}_{glb}$ and the local contrastive loss $\mathcal{L}_{lcl}$ to evaluate their significance.
We set their weights to $\{0.1, 1, 2\}$ and report the result in Table.~\ref{tab:loss_weight}.
It is demonstrated that setting the weights for $\mathcal{L}_{glb}$ to 1 and $\mathcal{L}_{lcl}$ to 0.1 achieves the best overall performance, yielding the highest dice score of 0.350, PG of 0.449, and CNR of 1.042.
Notably, increasing the weight for $\mathcal{L}_{lcl}$ to 1 or 2 leads to a reduction in the dice score and slight degradation in PG and CNR, suggesting that overemphasizing localization introduces diminishing returns. 
$\mathcal{L}_{glb}$ significantly influences segmentation performance, while localization loss has a lesser impact.

\begin{table}[h]
    \centering
    \caption{Loss weights versus performance.}
    \label{tab:loss_weight}
    \begin{tabular}{cc|ccc}
         \toprule $\mathcal{L}_{glb}$ & $\mathcal{L}_{lcl}$ & CNR$\uparrow$ & PG$\uparrow$ & Dice$\uparrow$  \\\midrule
          1 & 0.1 & 1.042 & 0.449 & 0.350 \\
          1 &   1 & 1.030 & 0.445 & 0.343  \\
          1 &   2 & 1.036 & 0.438 & 0.344 \\
          2 & 0.1 & 1.036 & 0.443 & 0.341 \\
          2 & 1   & 1.042 & 0.450 & 0.344 \\
          0.1 & 1 & 1.038 & 0.436 & 0.343 \\
          0.1 & 2 & 1.041 & 0.446 & 0.345
          \\\bottomrule
    \end{tabular}
\end{table}

\subsection{Robustness with other interpretability methods.}
Chefer et al.'s~\cite{chefer2021generic} method was chosen for its strong track record in VG tasks and alignment with the bi-modal nature of VLMs, which is often used in previous VG works in VPT, g, and g++.
To assess robustness, 
we implemented DAP with different methods 
on RSNA and COVID datasets 
with results in 
Tab.~\ref{tab:other_inter_method}.
It shows that GradCAM and SmoothGrad closely match the original performance, 
while LRP achieves similar scores to Self-EQ (CVPR2024). 
As such, DAP generalizes to other interpretability methods but we empirically found that ~\cite{chefer2021generic} gives optimal performance.

\begin{table}[h]
    \centering
    \caption{CNR scores of different interpretability methods.}
    \label{tab:other_inter_method}
    \resizebox{0.48\textwidth}{!}{
    \begin{tabular}{l|cccc|c}
             \toprule & Chefer et al[8]       & GradCAM & SmoothGrad & LRP   & Self-EQ  \\\midrule
        RSNA  & \textbf{1.630} & 1.462   & 1.507      & 0.973 & 1.075 \\
        COVID & \textbf{1.018} & 0.903   & 0.891      & 0.612 & 0.659
    \end{tabular}%
    }
\end{table}

\subsection{Prompt depth}

We investigate the impact of prompting at different layers within the vision encoder. Specifically, we evaluate prompting at the pixel space of the original image, the first half of the encoder, the full encoder, the second half, and exclusively at the last layer.
We report the result in Table.~\ref{tab:prompt_depth}.
The "last layer" setup achieves the best trade-off between segmentation accuracy and robustness, offering a strong balance between dice and CNR.
In contrast, configurations like "full" or "last half" slightly enhance CNR but compromise dice, underscoring the effectiveness of focusing on deeper features for balanced performance.
In contrast, the "first only" and "first half" configurations underperform.

\begin{table}[h]
    \centering
    \caption{Disease aware prompting layer depth versus performance.}
    \label{tab:prompt_depth}
    \begin{tabular}{r|ccc}
         \toprule layer & CNR$\uparrow$ & PG$\uparrow$ & Dice$\uparrow$  \\\midrule
          first layer & 1.037 & 0.443 & 0.341 \\
          first half & 1.039 & 0.442 & 0.343  \\
          full & 1.043 & 0.450 & 0.343  \\
          last half & 1.046 & 0.443 & 0.344  \\
          last layer & 1.042 & 0.449 & 0.350 \\\bottomrule
    \end{tabular}
\end{table}


\subsection{Fixed versus Varied text Prompts}
We investigate textual prompting strategies by exploring one fixed prompt, and multiple paraphrases per disease class. The settings range from 1 prompt per class to 20 and 50 prompts per class, with results summarized in Table.~\ref{tab:num_prompts}. The findings demonstrate that increasing the number of prompts per class enhances model performance, likely due to expanded vocabulary exposure, which improves the robustness of the text model.

\begin{table}[h]
    \centering
    \caption{Number of text prompts per class versus performance.}
    \label{tab:num_prompts}
    \begin{tabular}{c|ccc}
         \toprule \# prompts & CNR$\uparrow$ & PG$\uparrow$ & Dice$\uparrow$  \\\midrule
          1  & 0.991 & 0.417 & 0.334 \\
          10 & 1.037 & 0.443 & 0.344  \\
          20 & 1.036 & 0.448 & 0.349  \\
          50 & 1.042 & 0.449 & 0.350 \\\bottomrule
    \end{tabular}
\end{table}

\section{Few-shot Supervised finetuning performance}

We further evaluate the proposed approach in few-shot settings, starting with weakly-supervised training followed by 20-shot fine-tuning using ground truth dense labels.
As shown in Table.~\ref{tab:few_shot}, our proposed DAP achieves results under weakly-supervised settings that are on par with the 20-shot fine-tuned performance of competing methods, demonstrating its efficiency in learning meaningful representations with minimal supervision for visual grounding.

\begin{table*}[t]
\centering
\begin{tabular}{lc|cc|cc|cc}
\toprule
       &
       & \multicolumn{2}{c}{CNR$\uparrow$} & \multicolumn{2}{c}{PG$\uparrow$} & \multicolumn{2}{c}{Dice$\uparrow$} \\
\cmidrule(r){3-4} \cmidrule(r){5-6} \cmidrule(r){7-8}
       \textbf{Method} & Venue
       & weak & 20-shot & weak & 20-shot & weak & 20-shot \\
\midrule
g~\cite{shaharabany2022looking} & NIPS22     & 1.461 & 1.606   & 0.284 & 0.782   & 0.450 & 0.512   \\
g++~\cite{shaharabany2023similarity} & CVPR23    & 1.350 & 1.625   & 0.467 & 0.733   & 0.445 & 0.456   \\
VPT~\cite{lin2024visual} & ICASSP24    & 1.173 & 1.599   & 0.612 & 0.785   & 0.468 & 0.535   \\
\midrule
DAP & Proposed   & 1.630 & 1.741   & 0.747 & 0.816   & 0.474 & 0.551   \\
\bottomrule
\end{tabular}
\caption{Weakly-supervised and Few-shots finetuning performance on RSNA~\cite{shih2019augmenting} dataset.}
\label{tab:few_shot}
\end{table*}

\section{Text prompts construction}
We utilize GPT-4o~\cite{openai2023gpt4} to generate disease-centric descriptions for chest X-ray findings. The model is instructed to produce 50 distinct prompts for each disease, explicitly avoiding anatomical location details. The prompt we used is:
\begin{quote}
    For the disease \texttt{<disease name>}, provide a sentence to describe it on chest X-rays. Exclude any reference to anatomical locations and ensure findings are concise, medically accurate, and reflect a professional radiology reporting style.
\end{quote}

\section{Robustness to flawed $\Phi$}
\begin{figure}
    \centering
    \includegraphics[width=\linewidth]{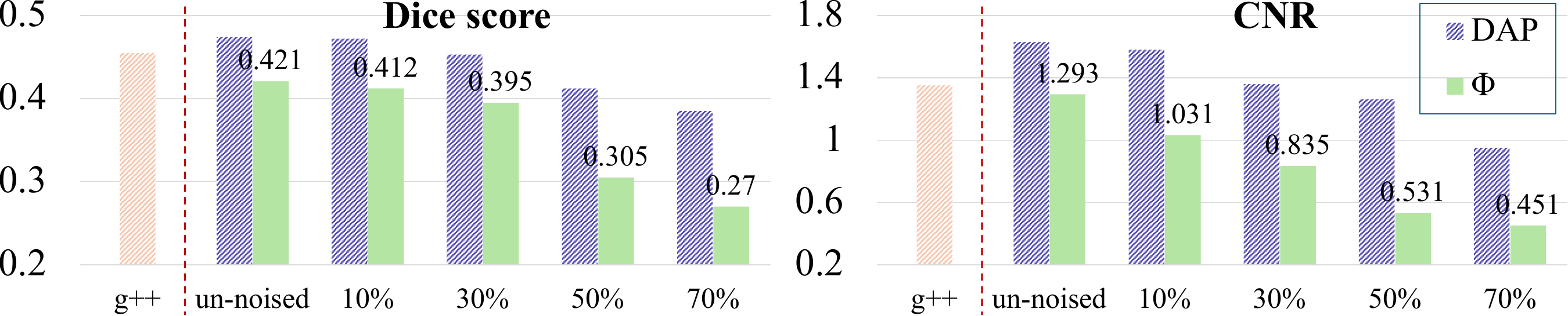}
    \caption{Dice score (\textit{left}) and CNR (\textit{right}) of DAP and $\Phi$ against the ground truth of RSNA dataset under different noise levels, compared to the strong baseline g++[44].}
    \label{fig:robustness}
\end{figure}

We inject noise into $\Phi$ and investigate \textit{how DAP degrades as noise increases}.
We first consider pixels of $\Phi$ with value $> 0.3$ as important.
Then, we flip top-$k\in\{10, 30, 50, 70\}\%$ of important pixels to $0$, and retrain the model to find \textit{how good should $\Phi$ be to benefit DAP}.
We conducted experiments on RSNA and plot the result in Fig.~\ref{fig:robustness}.
DAP's performance is indeed correlated with the quality of $\Phi$, which is lower than g++ when $\text{dice}(\Phi, \text{GT}) \leq 0.3$ at $k=50\%$.
Yet, we note that this is artificial scenario. In practice, $\Phi$ exhibits a reasonable localization
capability with $0.34$/$0.42$/$0.33$ dice score on MS-CXR/RSNA/Covid.

\section{Qualitative results}
\begin{figure*}[t]
\includegraphics[width=0.88\linewidth]{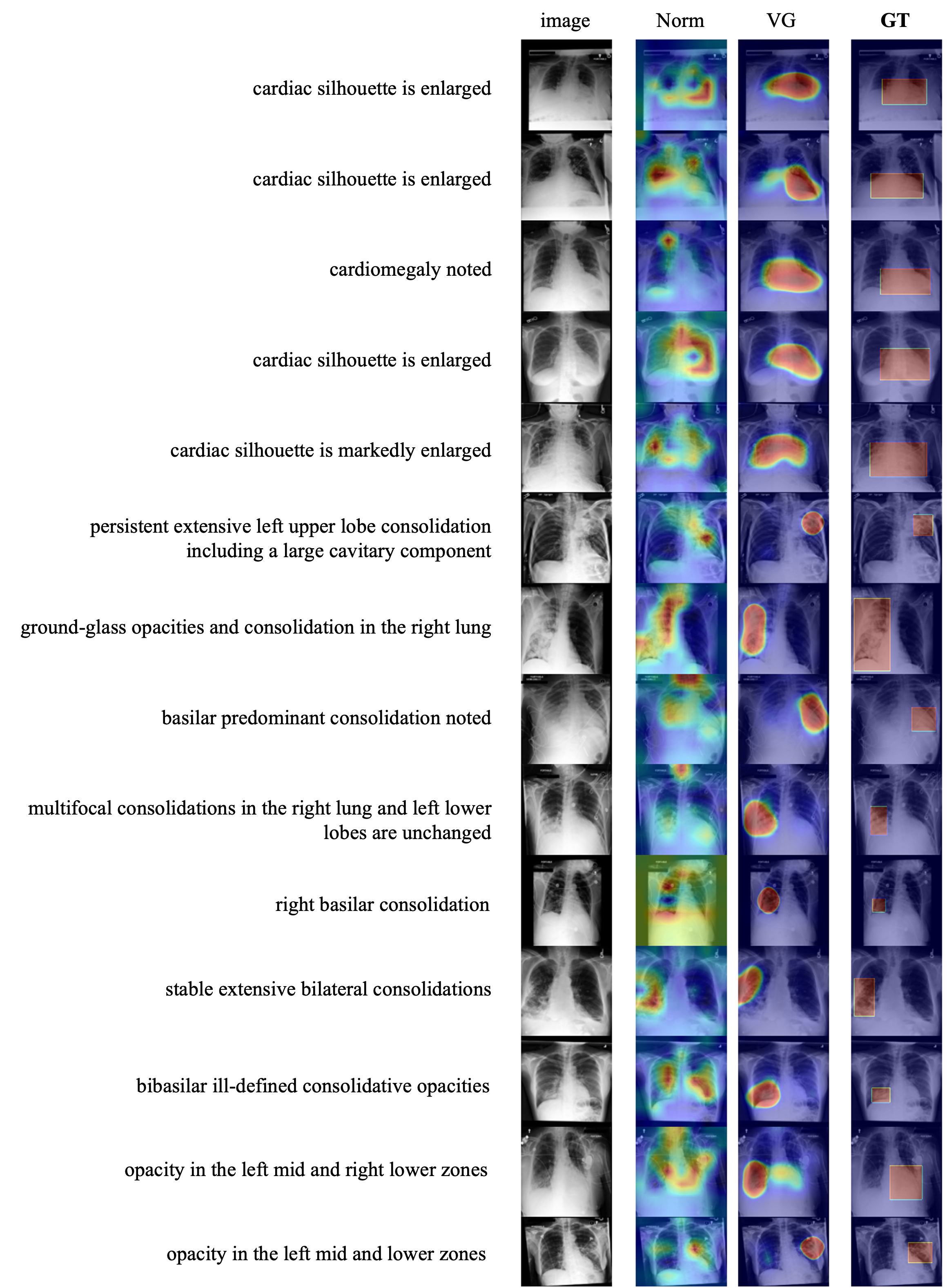}
    \caption{Qualitative results of DAP on MS-CXR~\cite{boecking2022making} dataset. }
    \label{fig:viz_more_1}
\end{figure*}

\begin{figure*}
\includegraphics[width=0.88\linewidth]{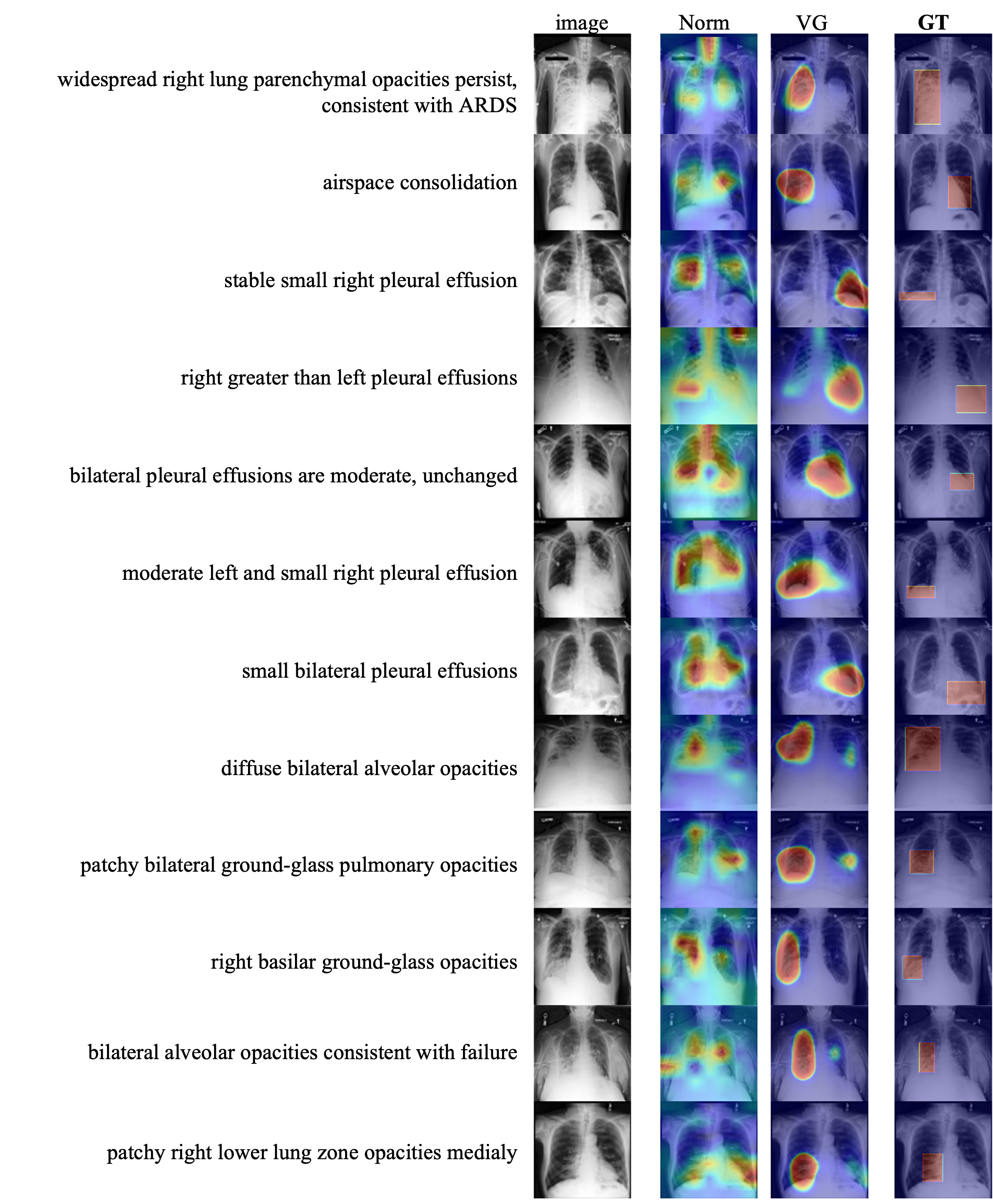}
    \caption{More qualitative results of DAP on MS-CXR~\cite{boecking2022making} dataset. }
    \label{fig:viz_more_2}
\end{figure*}

We present more qualitative results of our proposed DAP in Fig.~\ref{fig:viz_more_1} and Fig.~\ref{fig:viz_more_2}.

